\documentclass{article}

\usepackage{arxiv}

\usepackage[utf8]{inputenc} 
\usepackage[T1]{fontenc}    
\usepackage{hyperref}       
\usepackage{url}            
\usepackage{booktabs}       
\usepackage{amsfonts}       
\usepackage{nicefrac}       
\usepackage{microtype}      
\usepackage{lipsum}		
\usepackage{graphicx}
\usepackage{natbib}
\usepackage{color} 
\usepackage{amssymb} 
\usepackage{physics}
\usepackage{empheq}
\usepackage{caption}
\usepackage{subcaption}
\usepackage{breqn}
\usepackage{multirow}
\usepackage{pdfpages}
\usepackage{geometry}
\usepackage{doi}

\title{Energy-based General Sequential Episodic Memory Networks at the Adiabatic Limit}

\author{ Arjun Karuvally \\
	Manning College of Information and Computer Sciences \\
	University of Massachusetts Amherst \\
	Amherst MA 01007, USA \\
	\texttt{akaruvally@umass.edu} \\
	\And
	Terry J. Sejnowski \\
	Computational Neurobiology Laboratory \\
	The Salk Institute for Biological Studies \\
	La Jolla, CA 92037, USA \\
	\texttt{terry@snl.salk.edu} \\
	\AND
	Hava T. Siegelmann \\
	Manning College of Information and Computer Sciences \\
	University of Massachusetts Amherst \\
	Amherst MA 01007, USA \\
	\texttt{akaruvally@umass.edu} \\
}

\renewcommand{\shorttitle}{General Sequential Episodic Memory}

\hypersetup{
pdftitle={A template for the arxiv style},
pdfsubject={q-bio.NC, q-bio.QM},
pdfauthor={David S.~Hippocampus, Elias D.~Striatum},
pdfkeywords={First keyword, Second keyword, More},
}

\begin{document}
\maketitle

\begin{abstract}
	The General Associative Memory Model (GAMM) has a constant state-dependant energy surface that leads the output dynamics to fixed points, retrieving  single memories from a collection of memories that can be asynchronously preloaded. We introduce a new class of General Sequential Episodic Memory Models (GSEMM) that, in the adiabatic limit, exhibit temporally changing energy surface, leading to a series of meta-stable states that are sequential episodic memories. The dynamic energy surface is enabled by newly introduced asymmetric synapses with signal propagation delays in the network's hidden layer. We study the theoretical and empirical properties of two memory models from the GSEMM class, differing in their activation functions. LISEM has non-linearities in the feature layer, whereas DSEM has non-linearity in the hidden layer. In principle, DSEM has a storage capacity that grows exponentially with the number of neurons in the network. We introduce a learning rule for the synapses based on the energy minimization principle and show it can learn single memories and their sequential relationships online. This rule is similar to the Hebbian learning algorithm and Spike-Timing Dependent Plasticity (STDP), which describe conditions under which synapses between neurons change strength. Thus, GSEMM combines the static and dynamic properties of episodic memory under a single theoretical framework and bridges neuroscience, machine learning, and artificial intelligence.
\end{abstract}


\begin{figure*}[th!]
	\begin{subfigure}[b]{0.99\textwidth}
         \centering
		\includegraphics[scale=0.75]{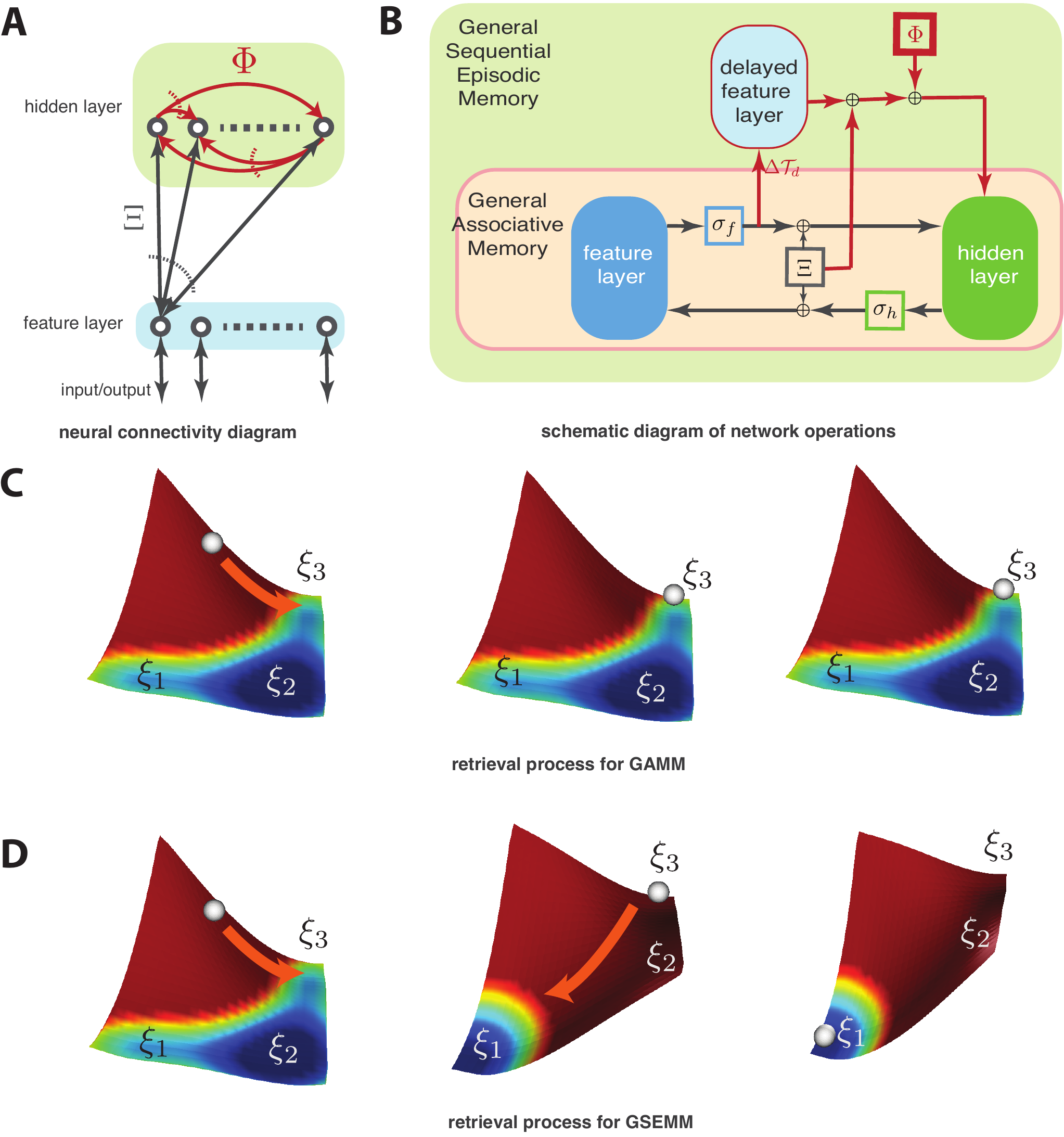}
     \end{subfigure}
\caption{ System architecture and schematic retrieval process for the General Sequential Episodic Memory Model(GSEMM) \textbf{A} The two-layer neural architecture with neural connectivity of GSEMM. The new delay-based synapses ($\Phi$) we introduced (shown in red) are directed connections between neurons in the hidden layer of GAMM. Dotted lines in the figure indicate one-to-many connections. \textbf{B} The schematic representation of how the network performs its computations. The new synapses create a delayed signal of the feature layer neurons, which is provided as input to the hidden layer. \textbf{C} The typical retrieval process for GAMM with three stored memories ($\xi_1$, $\xi_2$, $\xi_3$,). The energy surface is shown by the colormap with high energy denoted by red and low energy blue. The system (shown by the white ball) flows to the basin corresponding to the nearest attractor. Due to the energy surface's static nature, GAMM stays in a low-energy memory. Consequently, the system cannot retrieve more than one memory ($\xi_3$ in the figure). \textbf{D} Similar retrieval process for GSEMM. Unlike GAMM, the system changes the energy surface so that a new minimum is formed that connects to a sequentially related memory ($\xi_3 \rightarrow \xi_1$ in the figure). The dynamic nature of the energy surface allows the system to adapt to the new minimum under the condition that the changes in the energy surface are adiabatic to changes in state. This feature of GSEMM enables it to retrieve more than one memory. }
\label{fig:GEMM_energy_dynamics}
\end{figure*}

\section{Introduction}
Episodic memory refers to the conscious recollection of facts or subjective past experiences and forms an essential component of long-term memory \cite{Tulving2002EpisodicMF,Duff2019SemanticMA,Renoult2019FromKT}.
The recollection process may have both singleton and sequential characteristics.
Singleton retrieval is the associative recall of a single memory from a retrieval cue.
This memory could be the description of a particular object of interest or important dates of events.
Sequential retrieval leads to a recollection process that is not just a single memory but a chain of sequentially connected memories.
Episodic memory connects temporally related memories so that the retrieval process may consist of not just a single memory but sequential trajectories of these memories.
Memories organized into these trajectories are called \textit{episodes}.
Memories may come together in episodes allowing us to link and retrieve sometimes distinct and representationally unrelated memories.
The Sequential Episodic Memory (SEM) problem in Recurrent Neural Networks (RNNs) pertains to creating and manipulating these memories and their sequential relationships by encoding relevant information in some form in the synapses.

To date, associative recall based RNNs form the bulk of singleton episodic memory models.
Recent advances in energy-based associative memory showed how the memory recall property is universal for a class of symmetrically connected neural networks called the General Associative Memory Model (GAMM).
%
Associative memory models based on GAMM are models for singleton episodic memory since the retrieval process extracts a single memory associated with a retrieval cue.
One can imagine that if such an associative memory model stores sequential episodic memory, the episodes are stored as key-value pair mappings with time as the key and memory as the value.
However, there is a plethora of evidence to support the claim that neuronal populations encode episodic memory in the ordinal structure of their dynamic behavior \cite{Pastalkova2008InternallyGC, macdonald2011hippocampal, manns2007gradual, Mankin2012NeuronalCF}.
This evidence motivates the requirement for developing RNNs with sequential state transition characteristics.
The requirement is augmented by the many biological and machine learning systems \cite{Jones2007NaturalSE, PonceAlvarez2012DynamicsOC, Taghia2018UncoveringHB} that have sequential state transition characteristics.
In cognitive sciences, the systems underlie a range of processes related to working memory \cite{Taghia2018UncoveringHB}, perception \cite{Jones2007NaturalSE}, long-term decision making \cite{Wimmer2020EpisodicMR}, and inference and recall based on previous experiences \cite{Gupta2010HippocampalRI,Jin2014BasalGS}.

This paper explores a new class of General Sequential Episodic Memory Models (GSEMM) derived  by introducing delay-based synapses in the General Associative Memory Model. 
We define a slow-changing energy function that characterizes the dynamical nature of models from the GSEMM class as \textit{instantaneuos} fixed point attractor dynamics.
We study the SEM properties of two practical variants of GSEMM - Linear Interaction SEM (LISEM) and Dense SEM (DSEM), based on the type of interaction between memories in the energy function.
The memories have linear interactions in LISEM which are analogous to a model of sequentially activated memory \cite{Kleinfeld1986SequentialSG}.
In DSEM, we introduce non-linearity in the interactions between synapses \cite{Krotov2016DenseAM}.
We show that this introduction of non-linear interactions results in an exponential increase in SEM retrieval capacity over LISEM. 
Further, we use the energy paradigm to derive a learning rule for the synapses from the general theory so that DSEM can acquire new episodes online without preloading.
We show how the derived learning rules for the synapses are related to current biological plasticity rules: Hebbian, and Spike Timing Dependent Plasticity (STDP) \cite{Markam1997STDP}.
%
%
\section*{Energy-based Models}

The energy paradigm for memory was introduced by Hopfield \cite{Hopfield1982NeuralNA, Amari2004NeuralTO}, who defined energy as a quadratic function of the neural activity in symmetrically connected networks with binary model neurons.
A single memory is stored as a local minimum of the energy function.
The network dynamics converges to one of the local minima and retrieves a stable activity state representing a single episodic memory.
The Hopfield network model has subsequently been generalized along two directions.

The first direction focuses on memory capacity.
Capacity relates to the number of neurons required in the ensemble to store and retrieve a certain number of memories without corruption.
The capacity of the original Hopfield model was 14\% of the number of neurons, a small fraction of the number of neurons in the population\cite{McEliece1987TheCO, Folli2016OnTM, Amit1985SpinglassMO}.
A significant breakthrough in capacity came with the introduction of Dense Associative Memory \cite{Krotov2016DenseAM}, which introduced a polynomial non-linearity to separate the contribution of each memory to the energy minimum.
The non-linearity enabled the models to store more memories than the number of neurons (hence the term dense) with the caveat of introducing non-biological three-body interactions \cite{Krotov2021LargeAM}.
Further studies extended these ideas to continuous state spaces, and exponential memory capacity \cite{Widrich2020ModernHN}.
Currently, these models form the fundamental components of transformer architectures \cite{Vaswani2017AttentionIA, Ramsauer2021HopfieldNI} with high levels of performance on  large-scale natural language processing tasks \cite{Radford2018ImprovingLU, Devlin2019BERTPO} and computer vision \cite{Carion2020EndtoEndOD} tasks.
Recently, General Associative Memory Model (GAMM)  \cite{Krotov2021LargeAM} unified these advances in associative memory in a single theoretical framework.
GAMM succeeded in explaining the capacity improvements through a simple energy function that characterized the long-term behavior of these models just like its predecessors.
However, GAMM's state-parameterized constant energy surface restricts it to singleton episodic memories.

The second research direction focused on extending energy-based models to non-equilibrium dynamical conditions.
In contrast to memories in singleton episodic memory, memories in non-equilibrium models are meta-stable states in the dynamical evolution \cite{Rabinovich2008TransientCD, Durstewitz2008ComputationalSO, Camera2019CorticalCV}.
The non-equilibrium and sequential nature of the meta-stable states is an essential aspect of sequential episodic memory models.
Some of the first works to produce sequential meta-stable memory \cite{Kleinfeld1986SequentialSG, Sompolinsky1986TemporalAI} used a combination of symmetric interactions, asymmetric interactions, and delay signals to produce stable sequential activation of memory patterns.
However, these models required additional mechanisms to selectively raise the energies of states, which added complications to the use of the energy paradigm and showed the difficulty of reconciling  the static nature of the energy surface with  the dynamical nature of models required for sequential memory retrieval. 
One way to alleviate this difficulty is the introduction of stochasticity \cite{Miller2010StochasticTB, Jones2007NaturalSE} with sufficient noise to push the system's state beyond the basin of memory to another memory \cite{Miller2016ItinerancyBA, Braun2010AttractorsAN}.
Models developed along these directions relaxed the symmetric constraints on the neural interactions of Hopfield Networks, resulting in a rich repertoire of dynamics \cite{Asllani2018StructureAD, Orhan2020ImprovedMI}.
Theoretical proposals for meta-stable memory models used non-equilibrium landscapes where the energy function and a probability flux together determined the stability of memory states \cite{Yan2013NonequilibriumLT}.
In these models, stochasticity played a major role in determining the stability of meta-stable states.
Recent evidence from biology \cite{Howard2014, Rolls2019, Umbach2020TimeCI} has emphasized the importance of multiple timescales in SEM tasks.
Empirical models \cite{ Kurikawa2021MultipleTimescaleNN, Kurikawa2021TransitionsAM} also use multiple timescales to generate SEM.
However, current models do not take advantage of multiple timescales, and the SEM capacity of these models is only about 7\% of the number of neurons \cite{Kurikawa2021MultipleTimescaleNN}.

In GSEMM, we extend the energy paradigm of GAMM to the non-equilibrium case using two timescales to define the dynamic behavior of the model.
In the process, we discover mechanisms to significantly improve the sequential episodic memory capacity of non-equilibrium networks.
We derive learning rules that point to an intimate connection of local biological learning rules with a global energy minimization principle.

\section*{General Sequential Episodic Memory Model (GSEMM)}

We provide a mathematical description of GSEMM as a two layer system of interacting neurons organized according to the General Associative Memory Model (GAMM) \cite{Krotov2021LargeAM} with the addition of delay based intra-layer interactions between neurons in the hidden layer.
One of the layers is called the \textit{feature layer}.
This layer is mainly concerned with the input and output of the model.
There are \textit{no synaptic connections} between neurons in this layer.
The second layer is called the \textit{hidden layer}. 
The activity of neurons in this layer encodes abstract information about stored memories.
In contrast to the feature layer, the hidden layer neurons are connected using synapses that delay the signal from the feature layer.
These intra-hidden layer connections enable \textit{interactions between memories}.
In the most general case, there is no restriction on the nature (in terms of symmetry) of connections between neurons in this layer.
In addition to these intra-layer connections, the neurons in the two layers are connected through symmetric synaptic interactions.
%
%
The architecture for the model is shown in Figure \ref{fig:GEMM_energy_dynamics}.

%
We use common linear algebra notations and indexed notations to denote states and synapses in our model. 
We use mainly indexed notation but switch to matrix and vector notation wherever convenient.
We now provide the mathematical description of GSEMM. Let $(V_f)_i$ be the current through the $i^{\text{th}}$ neuron of the feature layer, $\sigma_f(V_f)$ be the activation function for the feature layer, $(V_h)_j$ be the current through the $j^{\text{th}}$ neuron of the hidden layer, $\sigma_h$ be the activation function for the hidden layer, and $(V_d)_i$ be the delayed feature neuron signal from the $i^{\text{th}}$ feature neuron.
The states $V_f, V_h, V_d$ evolve with characteristic timescales $\mathcal{T}_f, \mathcal{T}_h, \mathcal{T}_d$ respectively.
Let $\Xi_{i j}$ be the strength of the synaptic connection between the neuron $i$ in the feature layer to the neuron $j$ in the hidden layer, $\Phi_{k j}$ be the strength of the synaptic connection from the $k^{\text{th}}$ hidden neuron to the $j^{\text{th}}$ hidden neuron.
Similar to how memories are loaded in GAMM, each column of the matrix $\Xi$ stores individual memories.
We introduce two scalar parameters to control the strength of signals through the synapses. Let $\alpha_s, \alpha_c$ be the strength of signals through the synapses $\Xi$ and $\Phi$ respectively.
The governing dynamics are given by:
%
\begin{empheq}{equation}
\begin{dcases}
	\mathcal{T}_f \dv{(V_f)_i}{t} =& \sqrt{\alpha_s} \, \sum_{j=1}^{N_h} \Xi_{i j} \, (\sigma_h(V_h))_j - (V_f)_i \, , \\
	\mathcal{T}_h \dv{(V_h)_j}{t} =& \sqrt{\alpha_s} \sum_{i=1}^{N_f} \Xi_{i j} \, (\sigma_f(V_f))_i + \\ 
	                              &\alpha_c \sum_{k=1}^{N_h} \sum_{i}^{N_f} \Phi_{k j} \, \Xi_{i k} \, (V_{d})_i - (V_h)_j \, , \\
	\mathcal{T}_d \dv{(V_d)_i}{t} =& \, (\sigma_f(V_f))_i - (V_d)_i \, .
\end{dcases}
\end{empheq}
The dynamic evolution equations are analogous to GAMM \cite{Krotov2021LargeAM} except with the addition of intra-layer synapses $\Phi$, and two strength parameters $\alpha_s$ and $\alpha_c$.
The timescale $\mathcal{T}_d$ characterizes the timescale of delay and is assumed to be higher than the timescale of the feature and hidden layers.
The delay signal is obtained by applying a continuous convolution operator \cite{Kleinfeld1986SequentialSG} of the feature layer signal.
\begin{dmath}
	(V_d)_i = \frac{1}{\mathcal{T}_d} \int_{0}^\infty (\sigma_f(V_f(t-x)))_i \,\exp(-\frac{x}{\mathcal{T}_d}) dx .
\end{dmath}
We transformed the convolution operation to a dynamical state variable update $\dv{V_d}{t}$ to simplify the theoretical analysis of the system.

The General Associative Memory Model, which did not include intra-layer synapses in the hidden layer, has properties of associative memory.
This means that for certain conditions on the set of functions $\sigma_f$ and $\sigma_h$, the long-term behavior of the state of the feature layer neurons converged to one of the stored memories.
The crucial condition required for convergence is that the dynamical trajectory of the system followed an energy function with minima near the stored memory states.
The delay-based synapses we introduced enable the energy function to change with time, so the long-term behavior is not just a single memory but a sequence of related memories.

\section*{Energy Dynamics}

The energy dynamics of the system is analyzed by considering the new delay variable $V_d$ as a control parameter.
We show that for a delay signal $V_d$ that is changing \textit{sufficiently slowly} compared to $V_h$ and $V_f$, the energy function evaluated at the instantaneous state $V_d$ can still be used to characterize the dynamical nature of $V_f$ and $V_h$.
The term \textit{sufficiently slowly} means that $V_f$ and $V_h$ converge to their instantaneous attractor states before $V_d$ changes the energy surface.
To derive the energy function, we use two Lagrangian terms $L_f$ and $L_h$ for the feature and hidden neurons respectively \cite{Krotov2021LargeAM}, defined as
\begin{equation}
    \sigma_f(V_f) = \mathcal{J}(L_f)^\top, \,\, \text{and} \,\, \sigma_h(V_h) = \mathcal{J}(L_h)^\top \,.
\end{equation}
The new energy function (\textcolor{blue}{\textit{SI Appendix}: Energy Function for GSEMM}) for GSEMM is derived as.
\begin{dmath}
	E = \Bigg[ V_f^\top \, \sigma_f(V_f) - L_f\Bigg] + \Bigg[ V_h^\top \, \sigma_h(V_h) - L_h \Bigg] - \sqrt{\alpha_s} \, \Bigg[ \sigma_h(V_f) \, \Xi \, \sigma_h(V_h) \Bigg] - \alpha_c \Bigg[ V_d^\top \, \Xi \, \Phi \, \sigma_h(V_h) \Bigg] \, .
\end{dmath}
At this point, it is instructive to note that without the additional synapses $\Phi$ and strength parameter $\alpha_s = 1$, the system and the associated energy function reduce to GAMM energy with only singleton episodic memory.

In order to analyze how the dynamics of energy change with the introduction of delay based synapses, we take the time derivative of the GSEMM energy function along the dynamical trajectory of the system.
We assume the conditions of positive semi-definite Hessians of the Legrangian terms and bounded activation functions $\sigma_f$ and $\sigma_h$ \cite{Krotov2021LargeAM}.
It is to be noted that the full state description of the system consists of three vectors $V_f$, $V_d$, and $V_h$.
These states are grouped as a fast subsystem - $V_f$ and $V_h$, and a slow subsystem $V_d$.
The analysis becomes easier when we consider the slow subsystem as a control variable of the fast subsystem.
This allows the characterization of the state dynamics of the fast subsystem as instantaneous fixed point attractor dynamics modulated by input from the slow subsystem.

The dynamical evolution of the energy function after separating the slow and fast subsystems is given as (\textcolor{blue}{\textit{SI Appendix}: Energy Function Dynamics}),
\begin{dmath}
	\dv{E}{t} = F(\dv{V_f}{t}, \dv{V_h}{t}) + G(\dv{V_d}{t}) \, .
\end{dmath}
%
\begin{align}
\begin{split}
F(\dv{V_f}{t}, \dv{V_h}{t}) =& - \Bigg[ \mathcal{T}_f \left(\dv{V_f}{t}\right)^\top \, \mathcal{H}(L_f)  \, \dv{V_f}{t} + \\
                             & \mathcal{T}_h \left(\dv{V_h}{t}\right)^\top \,  \mathcal{H}(L_h) \,  \dv{V_h}{t} \Bigg] \, . \\
G(\dv{V_d}{t})=& - \alpha_c \Bigg[ \sigma_h(V_h)^\top  \, \Phi^\top  \, \Xi^\top \, \dv{V_d}{t} \Bigg] \, .
\end{split}
\end{align}
$F$ and $G$ separate the contributions of the two timescales - the fast ($\{\mathcal{T}_f, \mathcal{T}_h\}$) and slow ($\{ \mathcal{T}_d \}$).
It can be easily seen that among the two terms, only $G$ is affected by the timescale of the delay signal. 
Just like in GAMM, under the assumption of positive semi-definite Hessian of the Lagrangian and bounded energy, we get,
\begin{dmath}
	F(\dv{V_f}{t}, \dv{V_h}{t}) \leq 0 \, .
\end{dmath}
The inequality means that the fast subsystem can have two possible long-term behaviors when $F$ eventually converges to zero. 
One behavior is convergence to a single stable state corresponding to minima of the energy function leading to fixed point attractor dynamics.
The second possible behavior is when the system moves in an iso-energetic trajectory without convergence.
%
In this paper, we focus only on the case of the fixed point attractor behavior of the system.

Like in GAMM, the fixed point attractor behavior of the system acts to stabilize the dynamics on the energy surface such that the energy is non-increasing and convergent, but unlike GAMM, delay based synapses lead to another term $G$.
\begin{dmath}
G(\dv{V_d}{t}) = - \alpha_c \Bigg[ \sigma_h(V_h)^\top  \, \Phi^\top  \, \Xi^\top \,  \dv{V_d}{t} \Bigg] \, .
\end{dmath}
%
It may be difficult to specify the behavior of the system for any general choice of $\Phi$, $\sigma_h$, and $V_d$.
However, in the adiabatic limit of the slow subsystem (under the condition that $\mathcal{T}_d \gg 1$ and $\mathcal{T}_d \gg \mathcal{T}_f, \mathcal{T}_h$), the system can still exhibit a non-increasing energy function because $\dv{V_d}{t} \rightarrow 0$ in this limit and $G \rightarrow 0$.
This condition is especially true when analyzing the dynamic properties of the fast subsystem ($\dv{V_f}{t} \neq 0$ and $\dv{V_h}{t} \neq 0$) which is the property that seems to be relevant in dynamic memory models.
The delay signals thus have two functions.
The slow changing nature of the delay signal helps to stabilize the dynamics of the fast subsystem on the energy surface.
The second function is that the delay signal changes the energy surface to create new minima and destroy old minima.
In our numerical simulations, we consider high enough settings of $\mathcal{T}_d$ such that $V_d$ changes sufficiently slowly for the energy function to characterize the dynamics but not so high as to prevent the system from exhibiting state transitions in a reasonable time. 

\section*{Practical GSEMM variants}

The theory of GSEMM alone is not practical enough to be applicable in a sequence generation task as it does not specify the activation functions for each of the layers.
We derive two variants depending on the settings of activation functions and apply them to a sequential state generation task.
Analogous to how practical models are derived from GAMM, we consider the diabatic limit of hidden neurons for the two variants.
In the diabatic hidden neuron limit, 
\begin{dmath}
(V_h)_j = \sqrt{\alpha_s} \sum_{i=1}^{N_f} \Xi_{i j} \, (\sigma_f(V_f))_i + \alpha_c \sum_{k=1}^{N_h} \sum_{i}^{N_f} \Phi_{k j} \, \Xi_{i k} \, (V_{d})_i \, .
\end{dmath}
Substituting this in the dynamical evolution of feature neurons we get,
\begin{align}
\begin{split}
	\mathcal{T}_f \dv{(V_f)_i}{t} =& \sqrt{\alpha_s} \, \sum_{j=1}^{N_h} \Xi_{i j} \, \sigma_h(\Xi^\top \, \sigma_f(V_f) + \alpha_c \Phi^\top \, \Xi^\top \, V_d)_j\\ & 
	- (V_f)_i \, .
\end{split}
\end{align}
It can be seen from the dynamical evolution of feature neurons that depending on the settings of the two activation functions, the feature-hidden synapses may interact linearly with hidden-feature synapses and hidden-hidden synapses.
The two variants of the general theory are constructed based on the presence or absence of these inter-synapse interactions.
In the first variant, Linear Interaction SEM, the feature layer activation function is non-linear, and hidden layer activation is identity allowing linear interactions between synapses.
In the second variant, Dense SEM, the hidden layer activation function is non-linear which prevents linear interactions between synapses.

\section*{Linear Interaction SEM (LISEM)}

\begin{figure*}[th!]
	\begin{subfigure}[b]{0.99\textwidth}
         \centering
		\includegraphics[scale=0.3]{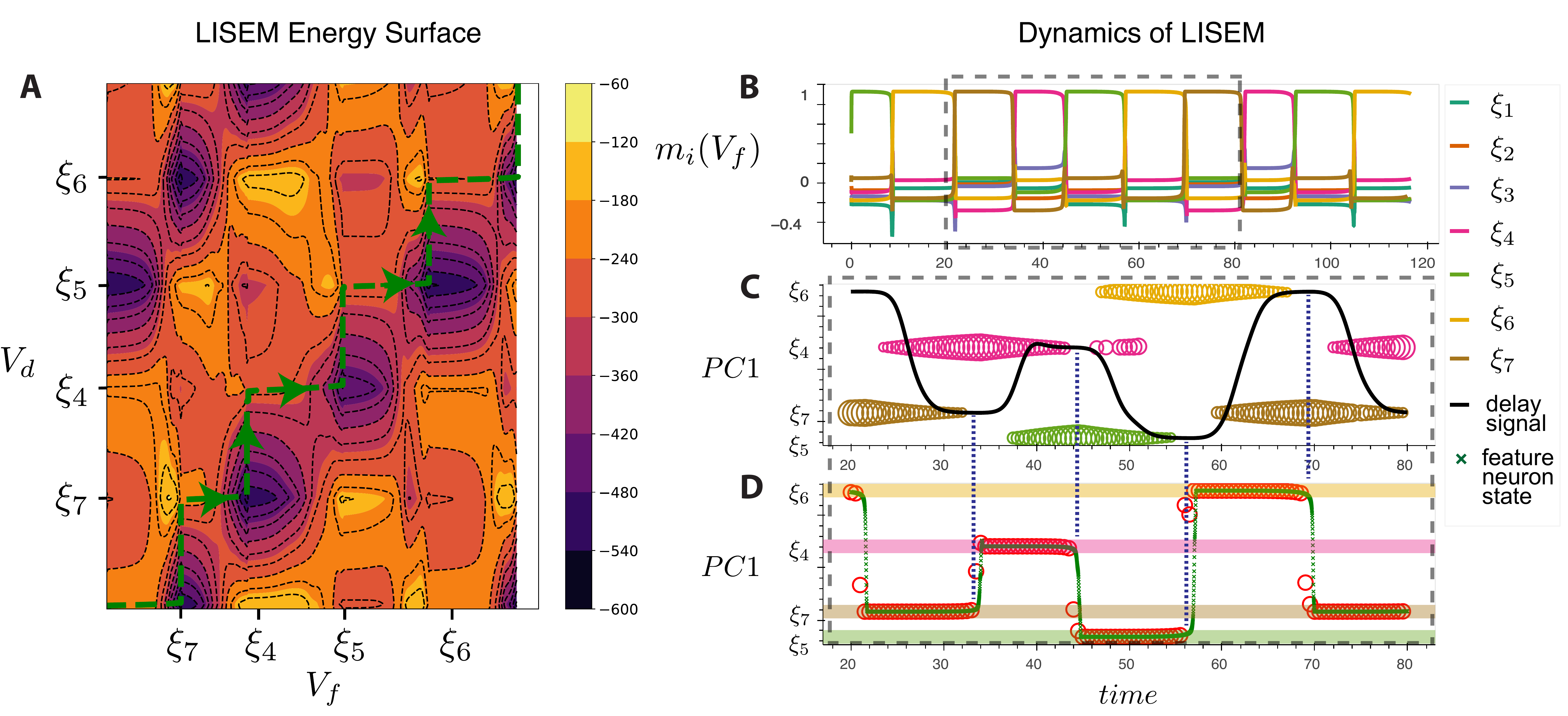}
     \end{subfigure}
\caption{ Energy dynamics of LISEM shows the instantaneous fixed point attractor behavior. \textbf{A} The energy surface of LISEM for the feature layer states $V_f$ (x-axis), and the delayed feature layer states $V_d$ (y-axis) visited by LISEM in one simulation during the time range $20 - 80$. The plot shows that $V_d$ changes the energy surface so that a new state for $V$ becomes a new basin of attraction (darker shade) while the previous basin increases energy. The trajectory (green line) of the system during retrieval shows the movement of $V_f$ to the \textit{instantaneous} minima of the energy surface. \textbf{B} The dynamics of $V_f$. The time range 20-80 over which energy behavior is plotted is shown in the gray dotted box. The y-axis denotes the overlap between the feature neuron state and each stored memory. \textbf{C} The first principle component ($PC1$) of $V_d$ and how it influences the fixed points near \textit{all} stored memories depicted by colored circles. The size of the circle is inversely proportional to the memories' energy. \textbf{D} The first principle component ($PC1$) of the dynamical evolution of the nearest fixed point (red circle) of the energy surface and the current state of the fast sub-system (green cross). The evolution shows how the system is attracted to the nearest fixed point from the current state of the energy surface at each point in time.  }
\label{fig:caseA_energy}
\end{figure*}

\begin{figure*}[t]
    \begin{subfigure}[b]{0.99\textwidth}
         \centering
		\includegraphics[scale=0.3]{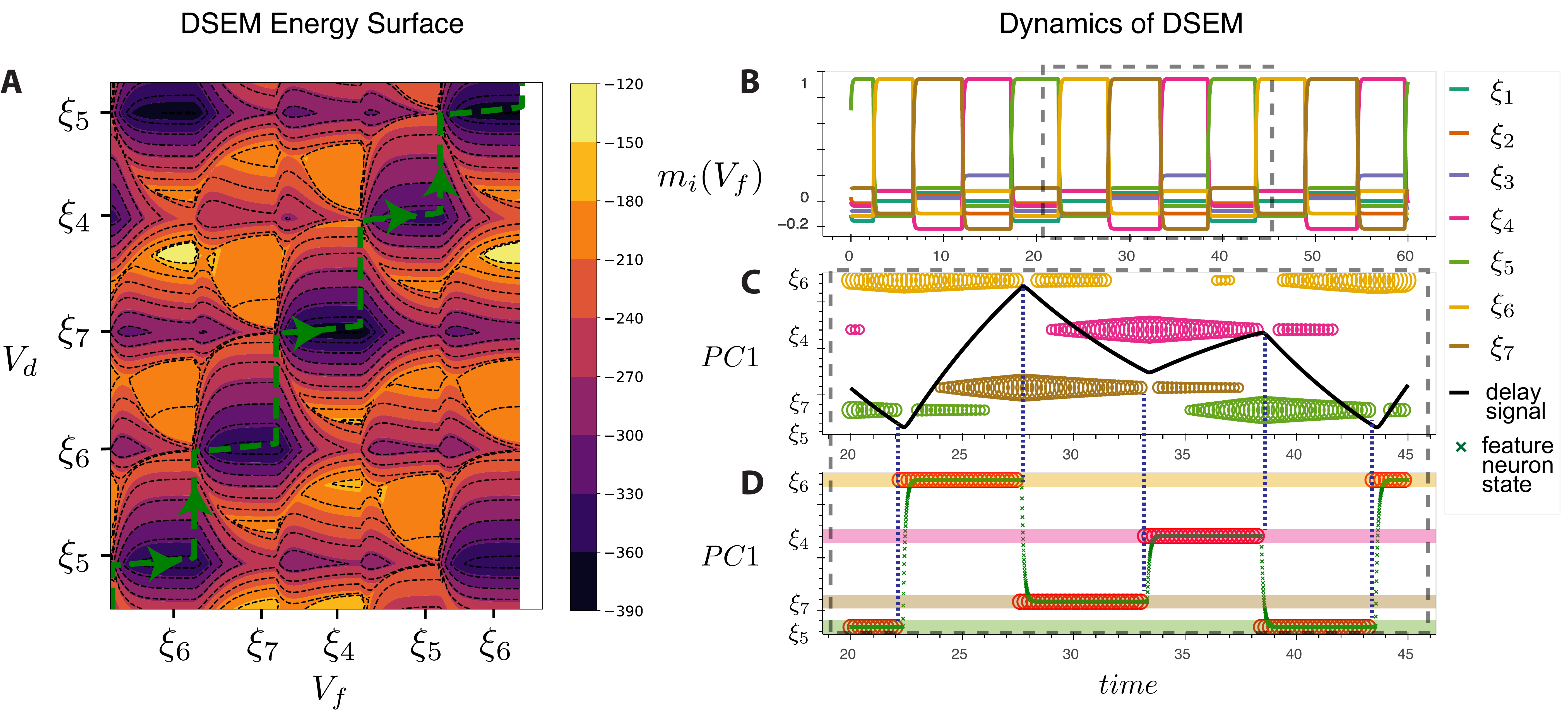}
     \end{subfigure}
\caption{ Energy dynamics showing the similarity and differences between DSEM and LISEM. \textbf{A} The energy surface of DSEM for the feature layer states $V_f$ (x-axis), and the delayed feature layer states $V_d$ (y-axis) visited by DSEM in one simulation during the time range $20 - 45$. Like LISEM, $V_d$ changes the energy surface such that a new state for $V$ becomes a new basin of attraction (darker shade) while the previous basin increases energy. The trajectory (green line) of the system during retrieval shows the movement of $V_f$ to the \textit{instantaneous} minima of the energy surface. \textbf{B} The dynamics of $V_f$. The time range of $20-45$ over which energy behavior is plotted is shown in the gray dotted box. The y-axis denotes the overlap between the feature neuron state and each stored memory. \textbf{C} The first principle component $(PC1)$ of $V_d$ and how it influences the fixed points near \textit{all} memories depicted by colored circles. The size of the circle is inversely proportional to the memories' energy. \textbf{D} The first principle component $(PC1)$ of the dynamical evolution of the \textit{nearest} fixed points (red circle) of the energy surface and the current state of the fast sub-system (green cross). The evolution shows how the system is attracted to the current fixed point of the energy surface at each point in time. Compared to LISEM, figures B, C, and D reveal some sharp transitions in the behavior of the delay signal and feature neuron states - shown by sharp changes in the black line of C. These sharp transitions suggest the faster convergence of DSEM compared to LISEM. }
\label{fig:caseB_energy}
\end{figure*}
LISEM is characterized by linear interactions between synapses.
This model closely resembles an RNN model with sequentially activated patterns explored previously in \cite{Kleinfeld1986SequentialSG}.
To analyze the dynamical properties of the model, we assume $N_h$ random binary vectors of dimension $N_f$ as memories preloaded in $\Xi$.
We assume a specific structure for $\Phi$ that gives rise to networks with sequential transitions:
\begin{dmath}
	\Phi = \frac{1}{\sqrt{\alpha_s}} \, G
\end{dmath}
where $G$ is the graph's adjacency matrix with $N_h$ nodes and directed edges that represent sequential relationships between memories in the stored episodes.
This structure for interneuron connections allows us to encode episodes with Markovian memory transitions in our network.
In the diabatic limit of hidden neuron activity, $\mathcal{T}_h \rightarrow 0$, identity activation for the hidden layer, $\sigma_h(V_h) = V_h$, and $\tanh$ activation in the feature layer, $\sigma_f(V_f) = \tanh(\gamma V_f)$, the governing dynamics reduce to:
\begin{empheq}{equation}
\begin{dcases}	
	\mathcal{T}_f \dv{(V_f)_i}{t} =& \alpha_s \sum_{j=1}^{N_h} \sum_{k=1}^{N_f} \Xi_{i k} \, \Xi_{j k} \, \tanh(\gamma \, (V_f)_k) + \\ 
	& \alpha_c \sum_{l=1}^{N_h} \sum_{j=1}^{N_h} \sum_{m=1}^{N_f} \Xi_{i j} \Phi_{l j} \Xi_{l m} (V_{d})_{m} \\
	& - (V_f)_i \, ,\\
	\mathcal{T}_d \dv{(V_d)_i}{t} =& \tanh(\gamma \, (V_f)_i) - (V_d)_i \, .
\end{dcases}
\end{empheq}


To elucidate how a robust recall of the next memory may be possible with the model, we analyze the dynamics of the energy function.
As discussed before, due to the dynamical nature of the system's long-term behavior and changing delay signal, the system no longer follows a \textit{global} energy function with minima near memories like in associative memory models.
Instead, the system follows the \textit{instantaneuos} minima of the energy function and goes from memory to memory via slow updates to the energy function.
We analyze the energy function of LISEM for the case where the memories are orthogonal column vectors of $\Xi$.
\begin{align}
\begin{split}
E_{\text{\tiny LISEM}} =& \Bigg[ V_f^\top \, \tanh(V_f) - \log(|\cosh(V_f)|) - \\ & \frac{\alpha_s}{2} \tanh(V_f)^\top \, \Xi \, \Xi^\top \, \tanh(V_f) \Bigg] \\ 
& + \Bigg[\alpha_c \, \tanh(V_f)^\top \, \Xi \, \Phi^\top \Xi^\top \, V_d \Bigg] \\ 
& + \Bigg[\frac{\alpha_c^2}{2 \alpha_s} V_d^\top \, \Xi \, \Phi \, \Phi^\top \, \Xi^\top \, V_d \Bigg] \, .
\end{split}
\end{align}
The energy function can be separated to three components as follows.
\begin{empheq}{equation}
\begin{dcases}
	E_{\text{\tiny LISEM}} =& E_{\text{\tiny assoc}} + E_{\text{\tiny seq}} + E_{\text{\tiny c}} \, .\\
	E_{\text{\tiny assoc}} =& \Bigg[ V_f^\top \, \tanh(V_f) - \log(|\cosh(V_f)|) \, .\\
	& - \frac{\alpha_s}{2} \tanh(V_f)^\top \, \Xi \, \Xi^\top \, \tanh(V_f) \Bigg] \, .\\
	E_{\text{\tiny seq}} =& \alpha_c \tanh(V_f)^\top \, \Xi \, \Phi^\top \Xi^\top \, V_d \, .\\
	E_c =& \frac{\alpha_c^2}{2 \, \alpha_s} V_d^\top \, \Xi \, \Phi \, \Phi^\top \, \Xi^\top \, V_d \, .
\end{dcases}
\end{empheq}
$E_{\text{\tiny assoc}}$ creates minima of $V_f$ near \textit{all} columns (memories) defined in $\Xi$.
This term is independent of $V_d$ and hence does not change over time.  
$E_{\text{\tiny c}}$ is independent of $V_f$ and just translates the energy surface.
Since energy functions are invariant under translation, the effect of this term can be safely ignored in the analysis.
Unlike $E_{\text{\tiny assoc}}$ and $E_{\text{\tiny c}}$, $E_{\text{\tiny seq}}$ is modulated by the the delay neurons $V_d$ with the strength $\alpha_c$.
Thus, depending on the state of $V_d$ and the parameter $\alpha_c$, the energy function creates different minima over time.
For the matrix $\Phi$ defined, the new minima are in the sequentially connected neighbor of the memory.


According to the theoretical analysis of the energy function dynamics of GSEMM, in the limit of slow changing signal $V_d$, the fast subsystem would follow the instantaneous minima of the energy function.
We validate this with simulation.
To show \textit{how} the state transition behavior is exhibited by LISEM, we plot  the energy function without $E_c$ and the state of the system as time progresses for a simulated episode of the system in Figure \ref{fig:caseA_energy}.
Since the state space is high dimensional and difficult to visualize, we show only the comparison between the energy of all states the system takes in $one$ simulation.
The figures also reveal how the system evolution follows the \textit{instantaneuous} minimum of the energy surface.

\begin{figure*}[t]
    \begin{subfigure}[b]{0.99\textwidth}
         \centering
		\includegraphics[scale=0.35]{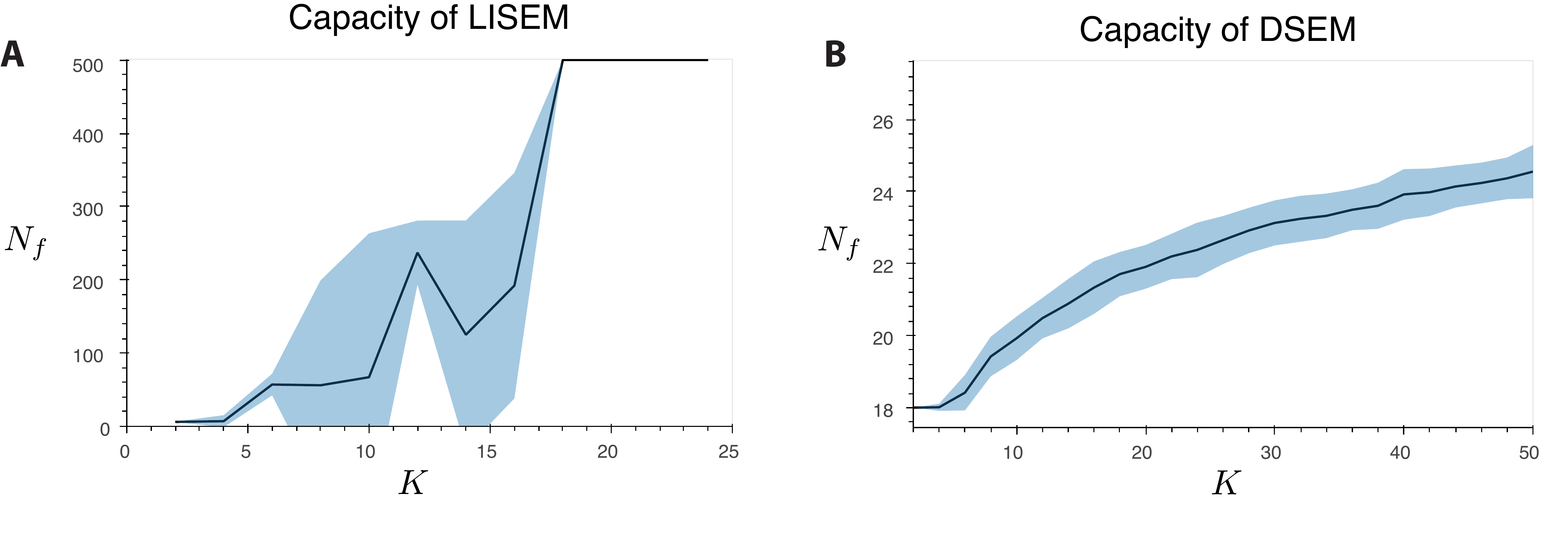}
     \end{subfigure}
\caption{ Capacity of LISEM and DSEM shows the exponential capacity of DSEM and the linear capacity of LISEM. The scaling relationship between the average number of neurons in the feature layer ($N_f$) of LISEM and DSEM to the number of memories that can be stored and retrieved ($K$). The blue-shaded region shows the regions within one standard deviation of the mean value of 100 trials. The number of feature neurons required for LISEM to encode episodes with a certain number of memories is shown to be significantly greater than DSEM. \textbf{A} LISEM seems to exhibit close to a linear relationship between $N_f$ and $K$ \textbf{B} DSEM shows a exponential relationship between $N_f$ and $K$.}
\label{fig:capacity}
\end{figure*}

\section*{Dense SEM (DSEM)}

The second variant of GSEMM is a model with structural improvements that greatly increase sequential episodic memory capacity to include longer sequence lengths $K$ exponential in the number of feature neurons $N_f$.
We use exponential interactions with contrastive normalization in the activation function of hidden layer neurons analogous to the Modern Hopfield Network \cite{Demircigil2017OnAM}.
\begin{equation*}
	\left( \sigma_h(V_h) \right)_i = \frac{\exp(\gamma (V_h)_i)}{\sum_{j=1}^{N_h} \exp(\gamma (V_h)_j)} \, .
\end{equation*} 
DSEM uses the identity activation for the feature layer $\sigma_f(V_f) = V_f$ leaving the non-linearity to the hidden layer activation function.
Under the diabatic conditions for the hidden layer, we get the dynamical equations for DSEM as:
\begin{empheq}{equation}
\begin{dcases}
	\mathcal{T}_f \dv{(V_f)_i}{t} =& \sqrt{\alpha_s} \, \sum_{j}^{N_h} \Xi_{i j} \, (\sigma_h(\sqrt{\alpha_s} \, \Xi^\top \, V_f + \alpha_c \, \Phi^\top \, \Xi^\top V_d))_j \\
	& - (V_f)_i \, . \\	
V_h =& \sqrt{\alpha_s} \, \Xi^\top \, V_h + \alpha_c \, \Phi^\top \, \Xi^\top V_d \, .\\
	\mathcal{T}_d \dv{(V_d)_i}{t} =& (V_f)_i - (V_d)_i \, .
\end{dcases}
\end{empheq}
Figure \ref{fig:caseB_energy} demonstrates that the local and global flow to attractors is similar to that of LISEM.
It is observed in the figure that the memory transitions are quicker than in LISEM which could be due to the rapid convergence rate of the associative memory system this model is based on \cite{Ramsauer2021HopfieldNI}.
The energy function of DSEM is
\begin{dmath}
	E_{\text{\tiny DSEM}} = \frac{1}{2} V_f^\top \, V_f + V_h^\top \, \sigma_h(V_h) - \text{logsumexp}(\exp(V_h)) - \sqrt{\alpha_s} \, V_f^\top \, \Xi \, \sigma_h(V_h) - \alpha_c V_d^\top \, \Xi \, \Phi^\top \, \boldsymbol{\sigma}(V_h) \, .
\end{dmath}
The energy dynamics shown in Figure \ref{fig:caseB_energy} also exhibit similar behavior to LISEM.
The momentary loss in stability of fixed points near memories that allow for state transition is observed clearly in the figure.

The crucial difference between LISEM and DSEM is the improved SEM capacity. 
Figure \ref{fig:capacity} compares LISEM and DSEM based on their SEM capacity, defined as the number of memories $K$ per sequential episode for a network with $N_f$ feature neurons. 
The required number of feature neurons is averaged over $100$ trials with different binary vectors encoding the memories.
To make the experiment computationally tractable, we set the maximum number of feature neurons at $500$ neurons.
The computational simulations suggest that DSEM is exponentially superior in capacity to LISEM, which stores only memories linear in the number of feature neurons.

\section*{Online Energy Learning for DSEM}

\begin{figure*}[t]
	\begin{subfigure}[b]{0.33\textwidth}
         \includegraphics[scale=0.36]{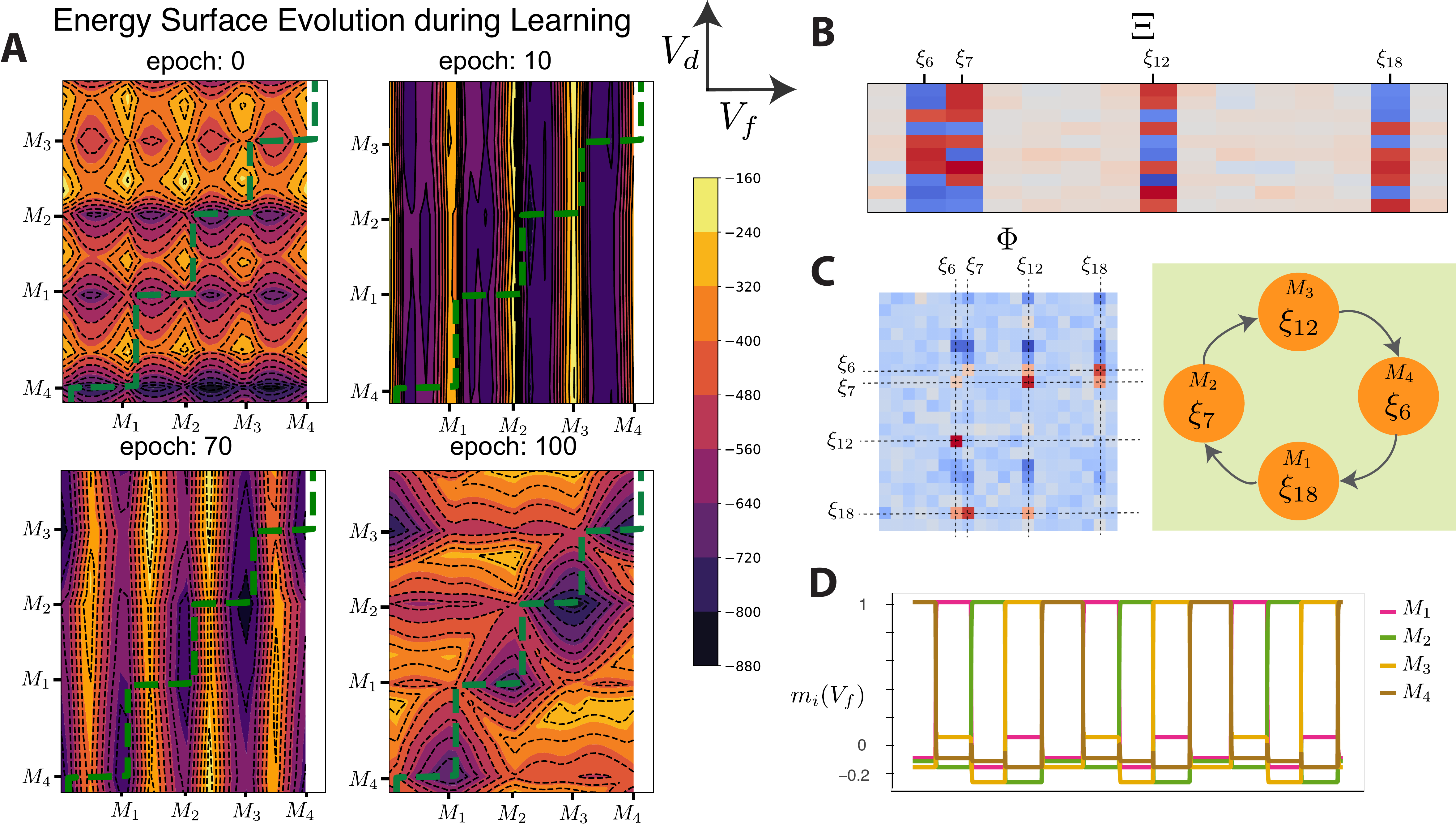}
     \end{subfigure}
\caption{ Energy and synaptic characteristics of DSEM during and after energy-based learning \textbf{A} The evolution of the energy surface during learning along the trajectory of DSEM (shown in green) when it is driven by the sequence of memories. At epoch 0, when no learning has occurred, the basins are created at states that don't correspond to the memories. As training progresses, in epoch 10, the emergence of large basins is observed that separate the memories, but the influence of delay on the energy surface is minimal. At epoch 70, the delayed feature neurons start to impact the energy surface. At the end of training at epoch 100, large basins are created that facilitate robust meta-stable memory transitions. \textbf{B} The structure of $\Xi$ after learning. The memories in the episode are consolidated in the synapses ($\xi_6, \xi_7, \xi_{12}, \xi_{18}$) observed as synapses with higher relative strengths. The plot is clipped for easier visualization of synapses. \textbf{C} The learned structure of $\Phi$ represents an adjacency matrix of the sequential relationship between stored memories. \textbf{D} The output of the learned network when a noisy representation of $\xi_{18}$ is provided as input.}
\label{fig:learning}
\end{figure*}

While Hopfield-like models assume preloaded memories by fixing weights rather than by learning, we propose here an online learning procedure that updates weights online to include more episodes. 
The online learning rule tunes the synaptic connections as the stimulus is provided as input to the model so that new sequential episodes can be learned by the model. 
The learning rule is derived from GSEMM using the following update rule for each synaptic connection $W = \{ \Xi, \Phi \}$.
\begin{dmath}
	\mathcal{T}^W_L \dv{W}{t} = - \pdv{E(x^{\text{target}})}{W} + \beta_c \pdv{E(x^{\text{current}})}{W} \, .
\end{dmath}
$\mathcal{T}^W_L$ is the characteristic learning rate for the synaptic connection $W$.
$x^{\text{current}}$ and $x^{\text{target}}$ are neuronal signals ($V_f, V_d, V_h$) estimated from the current and expected memory of the neurons respectively.
There are two different terms in the learning rule. 
The first term $-\dv{E(x^{\text{target}})}{W_{i j}}$ changes the model parameters such that the energy of the target memory decreases. 
This promotes the creation of attractor basins near the target memory.
The second term $\beta_c \dv{E(x^{\text{current}})}{W_{i j}}$ increases the energy of the current memory such that it destabilizes to flow to the target memory.
The learning rule is designed to mimic the expected dynamical behavior of the energy function when the system produces the required memory transitions.
We derive learning rules for the two synaptic interactions of DSEM based on our learning paradigm.
For $\Xi$ and $\Phi$, the learning rules are
\begin{dmath}
	\mathcal{T}_L^{\Xi} \dv{\Xi}{t} = \sqrt{\alpha_s} \Bigg[ V_f^{\text{target}} \sigma_h(V_h^{\text{target}})^{\top}  - V_f^{\text{current}} \sigma_h(V_h^{\text{current}})^{\top} \Bigg]
    + \alpha_c \, V_d \, \left( \sigma_h(V_h^{\text{target}})^{\top} - \sigma_h(V_h^{\text{current}})^{\top} \right) \, \Phi^\top \, .
\end{dmath}
\begin{dmath}
	\mathcal{T}_L^{\Phi} \dv{\Phi}{t} = \alpha_c \, \Xi^{\top} \, \left(   V_d \, \sigma_h(V_h^{\text{target}})^\top -  V_d \, \sigma_h(V_h^{\text{current}})^\top \right) \, .
\end{dmath}
The term $V_f^{\text{target/current}} \, \sigma_h(V_h^{\text{target/current}})^{\top}$ is the Hebbian learning rule between the \textit{feature neurons} and the \textit{hidden neurons}. 
Similarly, $V_d \, \sigma_h(V_h^{\text{target/current}})^\top$ is the Hebbian rule between the \textit{delayed feature neurons} and the \textit{hidden neurons}. 
However, since \textit{delayed feature neurons} store a delayed signal from the \textit{feature neurons}, this Hebbian term is actually the STDP learning rule between the \textit{feature neurons} and the \textit{hidden neurons}.
The STDP terms dominate the learning rule for $\Phi$, which suggests a connection between temporally aware STDP learning and the temporal nature of the information stored in $\Phi$.
STDP and Hebbian learning rules use local information on the activity of just the post and pre-synaptic neurons without considering global network computation.
This relationship between biological learning rules and our learning rule points to the vital role of local biological learning in global network energy minimization.

In Figure \ref{fig:learning}, we demonstrate the effectiveness of these learning rules in learning new memories along with their sequential relationships from input stimulus.
The synapses are initialized uniformly randomly so that no new memories are preloaded.
We use a 4-memory sequence as a sequential cyclical episode to learn - $M_1 \rightarrow M_2 \rightarrow M_3 \rightarrow M_4 \rightarrow M_1$.
After learning, Figure \ref{fig:learning} shows the test time behavior of the learned model and how the learned synapses organize such that the memory representations are stored in $\Xi$ and the sequential relationships between memories in $\Phi$.
The memories are consolidated in the feature hidden layer interaction $\Xi$ as $ M_1 - \xi_{18}$, $M_3 - \xi_7$, $M_3 - \xi_{12}$, and $M_4 - \xi_{6}$ where $\xi_{i}$ is a vector representing the strength of interactions between all the feature layer neurons and the $i^{\text{th}}$ hidden neuron.
%

\section*{Conclusion and Biological Relevance}
We introduced the General Sequential Episodic Memory Model that can encode memories and their sequential relationships.
Central to this capability is the slowly-changing energy surface controlled by the newly introduced delay-based synapses.
We showed how the energy surface's slow-changing nature helps the system to instantaneously follow the fixed points on the energy surface.
We studied two models from the GSEMM class. Linear Interaction Sequential Episodic Memory, with linear synaptic interactions, that is analogous to a popular sequential episodic memory model.
Dense Sequential Episodic Memory has non-linear synaptic interactions that exponentially improve episodic memory capacity.
We further proposed a learning rule for DSEM and showed how it is related to online versions of biological learning rules: Hebbian, and Spike-Timing Dependent Plasticity.

The generality of GSEMM theory could impact the future design and analysis of sequential episodic memory in both biological memory systems and machine learning.
%
The energy-based learning rule shows the role of biological learning rules that use local neuron information in minimizing the energy function of the global network.
Further research is needed to explore GSEMM and the connection between the energy paradigm and other aspects of neural networks in both neural systems and network models.
%
The scaling improvements of DSEM may be directly applied to solve problems requiring sequential memory with low overhead and high storage capacity.
Neuroscience and machine learning have much to learn from each other to improve our understanding of the dynamics of memory in intelligence.

\bibliographystyle{unsrtnat}
\bibliography{references}  

\section*{Supplemental Information - Materials and Methods}

We used the fourth order Range-Kutta numerical procedure with step size $0.01$ for numerical simulations. 
The output of the models is the state of their feature neurons and evaluated using the overlap of the feature neuron state with each memory in the system defined as $m_{i}(V_f) = (1/N_f) \, \sum_{j}^{N_f} (\xi_{i})_j \, (\sigma_f(V_f))_j$ where $\xi^{(i)}$ is the $i^{th}$ memory in the system.
Seven memories are encoded in each of the models with each memory in the system a random binary vector such that $\Pr[\xi^{(i)}_j = +1] = \Pr[\xi^{(i)}_j = -1] = 1/2$.
These memories are organized as 2 separate cyclical episodes: $\xi_1 \rightarrow \xi_2 \rightarrow \xi_3 \rightarrow \xi_1$ and $\xi_4 \rightarrow \xi_5 \rightarrow \xi_6 \rightarrow \xi_7 \rightarrow \xi_4$ with their sequential relationships stored as an adjacency matrix in $G$. 
Two key factors were considered when we used two episodes for evaluation.
One factor is to demonstrate the ability of the models to extract only the memories about the related stored episode, even in the presence of other episodes.
The second factor is that the successful generation of the stored episode requires long-term non-equilibrium behavior, meaning that the meta-stable states observed do not lead to an equilibrium ground state.

We used an iterative process to find fixed points of energy surface starting from some neuron state on the energy landscape, the state is updated to follow the direction of the energy slope till no more updates are possible, indicating convergence to a fixed point on the energy surface.
This fixed point is also one meta-stable point in the network dynamics.

\subsection*{LISEM}

We simulate LISEM with $N_f = 100, \gamma = 1.0, \alpha_s = 0.05, \alpha_c = 4.9, \mathcal{T}_f = 1.0,$ and $\mathcal{T}_d = 100.0$.
The output of the system is evaluated by considering the overlaps $m_{i}(V_f)$ of the state of feature neurons with $i^{\text{th}}$ preloaded memory of the system defined as $m_{i}(V_f) = (1/N_f) \, \sum_{j}^{N_f} (\xi_{i})_j \, (V_f)_j$.

\subsection*{DSEM}

We simulate DSEM with $N_f = 100, \gamma = 1.0, \alpha_s = 1.0, \alpha_c = 4.9, \mathcal{T}_f = 1.0$, and $\mathcal{T}_d = 100.0,$.

\subsection*{Online Energy Learning for DSEM}

The network synapses are randomly initialized with values from the range $[-1, 1]$. 
The sequence of memories is presented one after the other, with each memory supplied as input to $V_f$ for $4500$ timesteps. 
The learning algorithm is run with a learning timescale of $\mathcal{T}^{\Xi}_L = 6.2e5$, $\mathcal{T}^{\Phi}_L = 6.2e7$ and the model parameters $\alpha_c = 0.991$, $\beta_c = 0.621$, $\alpha_s = 1.0$. 

\section*{Supplemental Information - Proofs}

we give mathematical derivations we used to introduce essential concepts in the main text. 

\section*{Energy}

The most important aspect of the model we discussed is the energy function. We use the function to show the behavior of the system in the adiabatic case and compute instantaneuos attractors.

\subsection*{Energy Function for GSEMM} \label{SI:energy_function_derivation}

Here, we will derive the Energy function of GSEMM starting from a previously derived energy function used for associative memory.
Assume a signal $\mathcal{I}_h$ applied to the neurons in the hidden layer.
\begin{dmath}
	E = \Bigg[ V_f^\top \, \sigma_f(V_f) - L_f\Bigg] + \Bigg[ (V_h - \mathcal{I}_H)^\top \, \sigma_h(V_h) - L_h \Bigg] - \Bigg[ \sqrt{\alpha_s} \, \sigma_f(V_f) \, \Xi \, \sigma_h(V_h) \Bigg]
\end{dmath}
In our case, the input signal comes from the delay signal activity $V_d$ and is given as $\mathcal{I}_h = \alpha_c \Phi^\top \, \Xi^\top \, V_{d}$ from our governing dynamics.
Substituting this in the energy equation
\begin{dmath}
	E = \Bigg[ V_f^\top \, \sigma_f(V_f) - L_f\Bigg] + \Bigg[ (V_h - \alpha_c \Phi^\top \, \Xi^\top \, V_{d})^\top \, \sigma_h(V_h) - L_h \Bigg] - \Bigg[ \sqrt{\alpha_s} \, \sigma_f(V_f)^\top \, \Xi \, \sigma_h(V_h) \Bigg]
\end{dmath}
Expanding this equation, we get
\begin{dmath}
	E = \Bigg[ V_f^\top \, \sigma_f(V_f) - L_f\Bigg] + \Bigg[ V_h^\top \, \sigma_h(V_h) - L_h \Bigg] - \Bigg[ \sqrt{\alpha_s} \, \sigma_f(V_f)^\top \, \Xi \, \sigma_h(V_h) \Bigg] - \alpha_c \Bigg[ V_{d}^\top \, \Xi \, \Phi \sigma_h(V_h)  \Bigg]
\end{dmath}

\subsection*{Energy Function Dynamics} \label{SI:energy_function_dynamics}

To find how the energy function behaves along the dynamical trajectory of the system. Taking the derivative of the energy function with respect time

\begin{dmath}
	\dv{E}{t} = \Bigg[ V_f^\top \, \mathcal{J}(\sigma_f) \, \dv{V_f}{t} + \sigma_f(V_f)^\top \dv{V_f}{t} - \dv{L_f}{t} \Bigg] 
	+ \Bigg[ V_h^\top \mathcal{J}(\sigma_h)\, \dv{V_h}{t} + \sigma_h(V_h)^\top \, \dv{V_h}{t} - \dv{L_h}{t} \Bigg] 
	- \Bigg[ \sqrt{\alpha_s} \, \sigma_f(V_f)^\top \, \Xi \, \mathcal{J}(\sigma_h) \, \dv{V_h}{t} + \sqrt{\alpha_s} \, \sigma_h(V_h)^\top \, \Xi^\top \, \mathcal{J}(\sigma_f) \, \dv{V_f}{t} \Bigg] 
	- \alpha_c \, \dv{t}\Bigg[ V_{d}^\top \, \Xi \, \Phi \mathcal{J}(\sigma_h) \dv{V_h}{t} +  \sigma_h(V_h)^\top \, \Xi^\top \, \Phi^\top \dv{V_d}{t} \Bigg]
\end{dmath}
The derivatives of the legrangian terms can be converted as $\dv{L_f}{t} = \sigma_f(V_f)^\top \dv{V}{t}$ and $\dv{L_h}{t} = f(V_h)^\top \dv{V_h}{t}$. Substituting these.
\begin{dmath}
	\dv{E}{t} = \Bigg[ V_f^\top \, \mathcal{J}(\sigma_f) \, \dv{V_f}{t} 
	+ V_h^\top \mathcal{J}(\sigma_h)\, \dv{V_h}{t} \Bigg]
	- \Bigg[ \sqrt{\alpha_s} \, \sigma_f(V_f)^\top \, \Xi \, \mathcal{J} \, \dv{V_h}{t} + \sqrt{\alpha_s} \, \sigma_h(V_h)^\top \, \Xi^\top \, \mathcal{J}(\sigma_f) \, \dv{V_f}{t} \Bigg] 
	- \alpha_c \, \dv{t}\Bigg[ V_{d}^\top \, \Xi \, \Phi \mathcal{J}(\sigma_h) \dv{V_h}{t} +  \sigma_h(V_h)^\top \, \Xi^\top \, \Phi^\top \dv{V_d}{t} \Bigg]
\end{dmath}
Rearranging terms
\begin{dmath}
	\dv{E}{t} = - \Bigg[ \left(\sqrt{\alpha_s} \, \sigma_h(V_h)^\top \, \Xi^\top - V_f^\top \right) \, \mathcal{J}(\sigma_f) \, \dv{V_f}{t} 
 + \left( \sqrt{\alpha_s} \, \sigma_f(V_f)^\top \, \Xi + \alpha_c V_{d}^\top \, \Xi \, \Phi - V_h^\top \right) \mathcal{J}(\sigma_h)\, \dv{V_h}{t} \Bigg]  
	- \alpha_c \Bigg[ \sigma_h(V_h)^\top \, \Xi^\top \, \Phi^\top \dv{V_d}{t} \Bigg]
\end{dmath}
Substituting from dynamical equations
\begin{dmath}
	\dv{E}{t} = - \Bigg[ \mathcal{T}_f \, \dv{V_f}{t}^\top \, \mathcal{H}(L_f) \, \dv{V_f}{t} 
 + \mathcal{T}_h \dv{V_h}{t}^\top \mathcal{H}(L_h)\, \dv{V_h}{t} \Bigg]  
	- \Bigg[ \sigma_h(V_h)^\top \, \Xi^\top \, \Phi^\top \dv{V_d}{t} \Bigg]
\end{dmath}

\section*{Energy based learning}

In the paper, we discuss how the energy based learning connects to some well known biological learning rules. In this section, we derive the relations we used using the new energy function.

\subsection*{Learning rule for $\Xi$}

We use the following rule to make changes to $\Xi$.
\begin{dmath}
	\mathcal{T}_L^{\Xi} \dv{\Xi}{t} = - \pdv{E(V_f^{\text{target}})}{\Xi} + \pdv{E(V_f^{\text{current}})}{\Xi}
\end{dmath}
\begin{dmath}
	\mathcal{T}_L^{\Xi} \dv{\Xi}{t} = \Bigg[ \sqrt{\alpha_s} \, V_f^{\text{target}} \sigma_h(V_h^{\text{target}})^{\top} + \alpha_c V_d \, (\Phi \sigma_h(V_h^{\text{target}}))^{\top} \Bigg] 
    - \Bigg[ \sqrt{\alpha_s} \, V_f^{\text{current}} \sigma_h(V_h^{\text{current}})^{\top} + \alpha_c V_d \, (\Phi \sigma_h(V_h^{\text{current}}))^{\top} \Bigg]
\end{dmath}
\begin{dmath}
	\mathcal{T}_L^{\Xi} \dv{\Xi}{t} = \sqrt{\alpha_s} \Bigg[ V_f^{\text{target}} \sigma_h(V_h^{\text{target}})^{\top}  - V_f^{\text{current}} \sigma_h(V_h^{\text{current}})^{\top} \Bigg]
    + \alpha_c \, V_d \, \left( \sigma_h(V_h^{\text{target}})^{\top} - \sigma_h(V_h^{\text{current}})^{\top} \right) \, \Phi^\top
\end{dmath}

\subsection*{Learning rule for $\Phi$}

We use the following rule to make changes to $\Phi$.
\begin{dmath}
	\mathcal{T}_L^{\Phi} \dv{\Phi}{t} = - \pdv{E(V_f^{\text{target}})}{\Phi} + \pdv{E(V_f^{\text{current}})}{\Phi}
\end{dmath}
\begin{dmath}
	\mathcal{T}_L^{\Phi} \dv{\Phi}{t} = \Bigg[ \alpha_c \Xi^{\top} \, V_d \,  \sigma_h(V_h^{\text{target}})^\top \Bigg] - \Bigg[ \alpha_c \Xi^{\top} \, V_d \,  \sigma_h(V_h^{\text{current}})^\top \Bigg]
\end{dmath}
\begin{dmath}
	\mathcal{T}_L^{\Phi} \dv{\Phi}{t} = \alpha_c \, \Xi^{\top} \, \left(   V_d \, \sigma_h(V_h^{\text{target}})^\top -  V_d \, \sigma_h(V_h^{\text{current}})^\top \right)
\end{dmath}

\end{document}